\documentclass[letterpaper]{article} 
\usepackage{aaai2026}  
\usepackage{times}  
\usepackage{helvet}  
\usepackage{courier}  
\usepackage[hyphens]{url}  
\usepackage{graphicx} 
\usepackage{natbib}  
\usepackage{caption} 
\usepackage{algorithm}
\usepackage{algorithmic}

\usepackage{latexsym}
\usepackage{multirow}
\usepackage{booktabs}
\usepackage{amsmath}
\usepackage{enumitem}
\usepackage{listings}
\usepackage{xcolor}
\usepackage{amssymb}
\usepackage{tabularx}

\usepackage{newfloat}
\usepackage{listings}
\DeclareCaptionStyle{ruled}{labelfont=normalfont,labelsep=colon,strut=off} 
\floatstyle{ruled}
\newfloat{listing}{tb}{lst}{}
\floatname{listing}{Listing}
\title{GranAlign: Granularity-Aware Alignment Framework for Zero-Shot Video Moment Retrieval}
\author{
    Mingyu Jeon\textsuperscript{\rm 1},
    Sunjae Yoon\textsuperscript{\rm 2},
    Jonghee Kim\textsuperscript{\rm 3},
    Junyeoung Kim\textsuperscript{\rm 1}
}
\affiliations{
    \textsuperscript{\rm 1}Department of Artificial Intelligence, Chung-Ang University\\
    \textsuperscript{\rm 2} Korea Advanced Institute of Science and Technology (KAIST)\\
    \textsuperscript{\rm 3}Electronics and Telecommunications Research Institute (ETRI)\\

    \{smart2557, junyeongkim\}@cau.ac.kr, sunjae.yoon@kaist.ac.kr, jhkim27@etri.re.kr
}

\begin{document}

\maketitle


\begin{abstract}
Zero-shot video moment retrieval (ZVMR) is the task of localizing a temporal moment within an untrimmed video using a natural language query without relying on task-specific training data. 
The primary challenge in this setting lies in the mismatch in semantic granularity between textual queries and visual content.
%
Previous studies in ZVMR have attempted to achieve alignment by leveraging high-quality pre-trained knowledge that represents video and language in a joint space.
%
However, these approaches failed to balance the semantic granularity between the pre-trained knowledge provided by each modality for a given scene. 
As a result, despite the high quality of each modality’s representations, the mismatch in granularity led to inaccurate retrieval.
In this paper, we propose a training-free framework, called \textit{\textbf{Gran}ularity-\textbf{A}ware A\textbf{lign}ment (GranAlign)}, that bridges this gap between coarse and fine semantic representations.
Our approach introduces two complementary techniques: granularity-based query rewriting to generate varied semantic granularities, and query-aware caption generation to embed query intent into video content. By pairing multi-level queries with both query-agnostic and query-aware captions, we effectively resolve semantic mismatches. As a result, our method sets a new state-of-the-art across all three major benchmarks (QVHighlights, Charades-STA, ActivityNet-Captions), with a notable 3.23\% mAP@avg improvement on the challenging QVHighlights dataset.
\end{abstract}


%
%
%
%

\section{Introduction}
Video Moment Retrieval (VMR)--the task of localizing a segment from a video via a language query--is crucial for efficient video understanding. However, traditional supervised approaches are fundamentally limited by their reliance on large, costly annotated datasets. To overcome this dependency, zero-shot VMR (ZVMR) has emerged as a powerful and successful paradigm, with its feasibility greatly enhanced by recent advances in Vision-Language Models (VLMs) and Large Language Models (LLMs).
%

Despite its success, the ZVMR paradigm faces a new, fundamental challenge: the `Granularity Mismatch' between the language query and the visual content. 
%
\begin{figure}[htb!]
\centerline{\includegraphics[width=1\columnwidth]{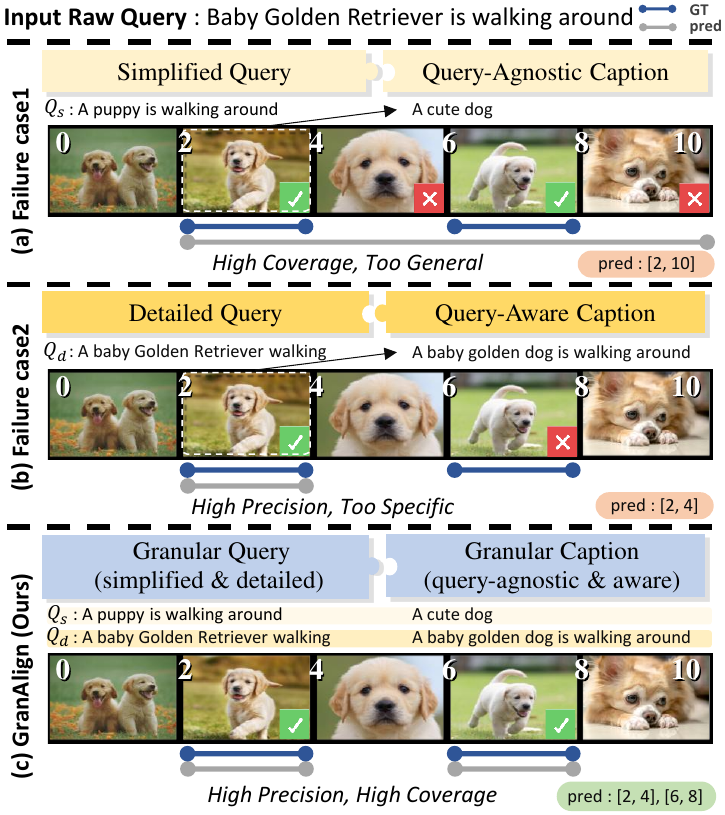}}
\caption{ An illustration of how GranAlign resolves the `Granularity Mismatch'. The (a) and (b) fail to localize all moments due to a mismatch in semantic granularity, resulting in low precision and low coverage, respectively. In contrast, our proposed GranAlign (c) integrates both granular levels, achieving both high precision and high coverage to correctly localize all target moments.\textbf{ Takeaway:} GranAlign overcomes the core `Granularity Mismatch' by synergizing the high-recall simple path with the high-precision detailed path.
}
\label{fig1}
\end{figure}
%
%
%
\begin{figure}[htb!]
\centerline{\includegraphics[width=1\columnwidth]{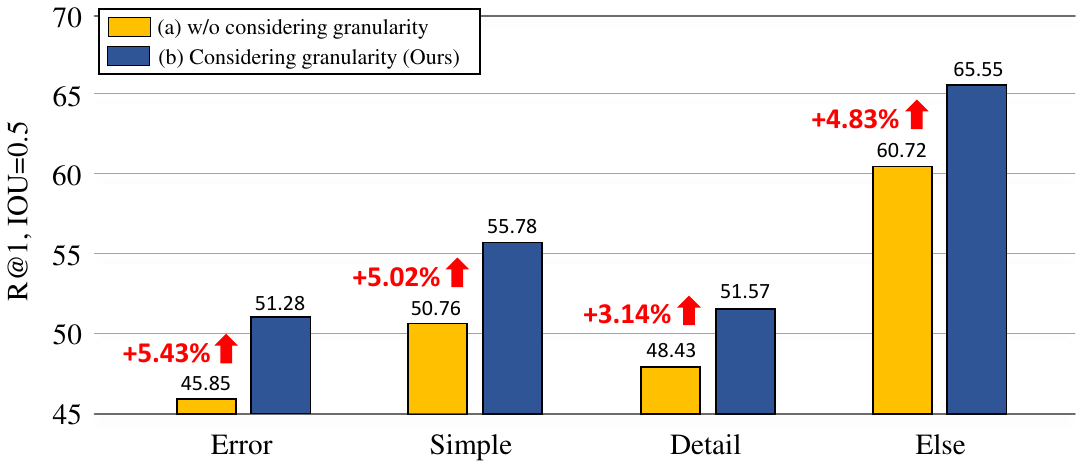}}
\caption{Comparison of our method (b) against a baseline (a) (without considering the granularity of query and video content) on the QVHighlights validation set, categorized by query type (see Section 4.3 for categorization criteria). 
}
\label{fig2}
\end{figure}
This issue arises because a user might describe the same event at varying levels of detail, from a general phrase like ``a cute dog" to a specific one like ``a baby Golden Retriever is walking around." As illustrated in Figure~\ref{fig1}, this mismatch creates an inevitable trade-off: a general query may achieve high coverage but lacks the precision to pinpoint the exact moment, while a specific query offers high precision but often suffers from poor coverage if subtle details do not perfectly align. 
%
This limitation is not merely anecdotal. Our quantitative analysis in Figure~\ref{fig2}, where performance is broken down by query type (Error, Simple, Detail, and Else), reveals a significant performance gap. This highlights the inability of current approaches to effectively handle varying levels of granularity. Prevailing methods, even when armed with high-quality pre-trained knowledge in a joint space, fundamentally fail to resolve this trade-off because they lack an explicit mechanism for granularity-aware alignment.

\begin{figure}[htb!]
\centerline{\includegraphics[width=1\columnwidth]{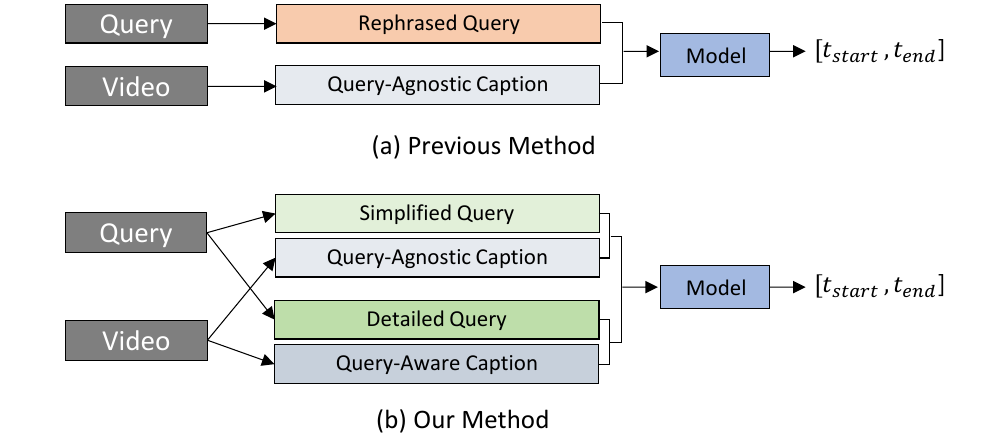}}
\caption{Conceptual Framework Comparison. (a) Previous methods typically adopt a single-path approach, reformulating the query. (b) Our method, GranAlign, employs a dual-path framework, decomposing the query into Simplified and Detailed versions and aligning them with Query-Agnostic and Query-Aware captions, respectively.
}
\label{fig3}
\end{figure}

The root cause of this problem is that existing methods treat queries monolithically. An intuitive solution might be to expand the original query with various rephrases, a strategy employed by several prior works as depicted in Figure~\ref{fig3} (a). However, this ``one-size-fits-all" approach is inherently flawed. Even a diverse set of rewritten queries, if confined to a single level of granularity, cannot simultaneously embody the broad scope needed for high recall and the fine-grained detail required for high precision. This approach typically results in a compromise that excels at neither, failing to adapt to the diverse semantic complexity across different query-video pairs. This single-pathway reasoning is the core bottleneck preventing robust and accurate retrieval.

To address these limitations, we introduce Granularity-Aware Alignment (GranAlign), a novel, training-free framework that models and aligns queries and video content at complementary levels of granularity. As conceptually illustrated in Figure~\ref{fig3} (b), GranAlign abandons the single-pathway design. Instead, on the query side, it leverages an LLM~\cite{llama3} to reformulate the original query into two distinct paths: a simplified query capturing the core intent for broad coverage, and a detailed query preserving specific nuances for precision. Correspondingly, on the video side, a VLM~\cite{qwen2.5vl} generates a general, query-agnostic caption and a focused, query-aware caption. By aligning these pairs by granularity--simplified query with query-agnostic caption and detailed query with query-aware caption--GranAlign synergizes the high-recall capability of the general path with the high-precision of the specific path, leading to a more robust alignment and superior retrieval accuracy across multiple challenging benchmarks.
%

\section{Related works}
\subsection{Zero-Shot Video Moment Retrieval}

Zero-shot Video Moment Retrieval (ZVMR) aims to localize relevant moments in a video using pretrained models without task-specific training. 
Prior work has achieved strong performance on the QVHighlights dataset using off-the-shelf vision-language models and BLIP-2-based~\cite{blip2} approaches~\cite{Diwan,wattasseril}. 
However, these methods often rely on coarse frame-level matching or restricted segment-based proposals, limiting their ability to achieve fine-grained temporal alignment and precise semantic grounding.
Recent advances such as Moment-GPT~\cite{Moment-GPT} leverage large language models (LLMs) to rephrase queries and use Video-ChatGPT~\cite{VideoChatGPT} to score candidate moments, achieving state-of-the-art performance.
Despite its advanced architecture, Moment-GPT still struggles with accurate semantic alignment between the rewritten queries and visual content, which hinders precise moment localization.
In the next subsection, we analyze this alignment issue through the lens of semantic granularity.


\begin{figure*}[htb!]
\centerline{\includegraphics[width=1\textwidth]{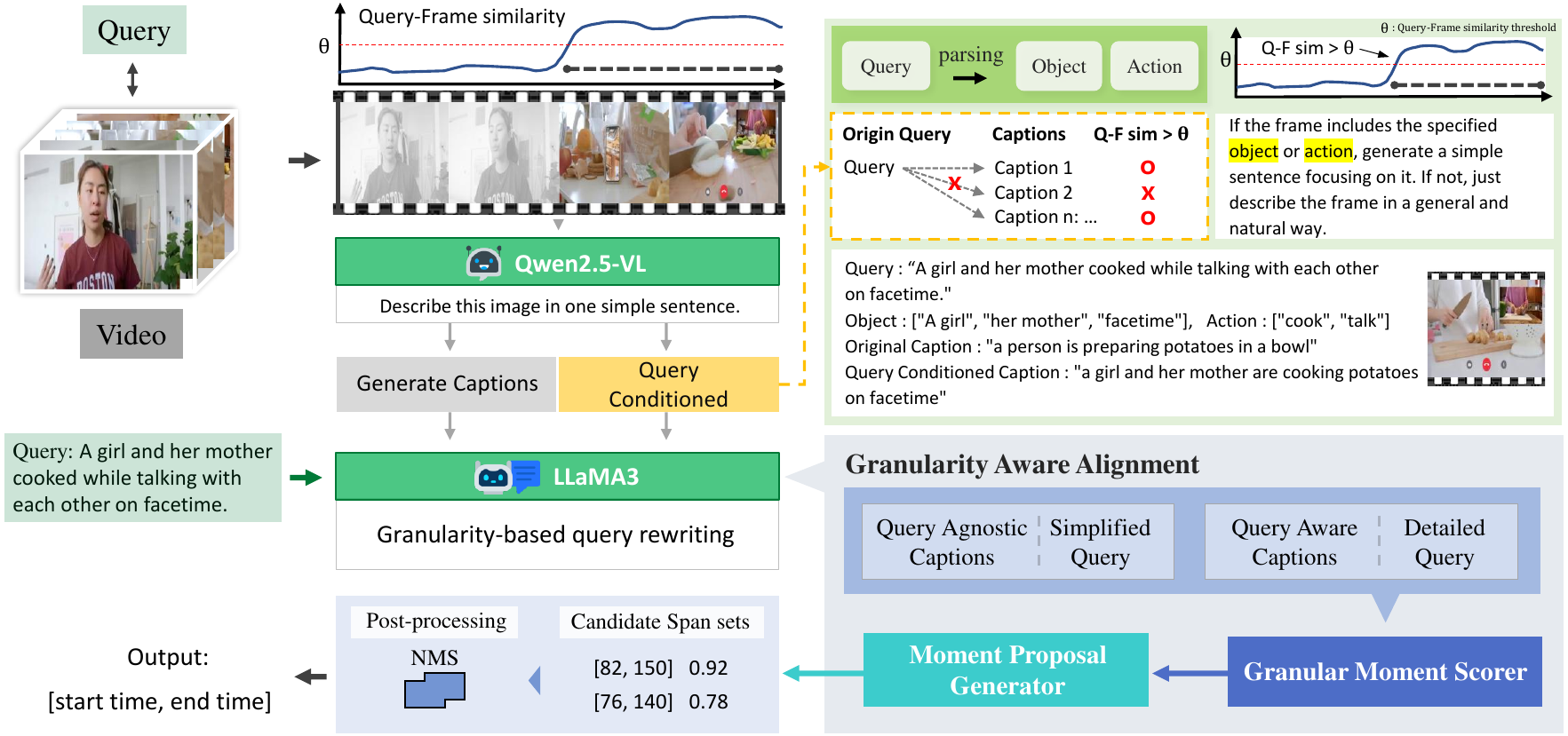}}
\caption{Overview of the GranAlign Framework. 
In Granularity-Aware Alignment (Sec.\ref{sec: scoring}), the input query is rewritten at two semantic granularities (simplified/detailed) and matched with either query-agnostic or query-aware captions (captions generated with the query as context) to obtain a \textit{Moment Score} for each video segment.  
These scores drive the Moment Proposal Generator and NMS stage, after which the post-ranking module produces the final prediction (Sec.\ref{sec: proposal}). \textbf{Takeaway:} GranAlign effectively tackles the ZVMR task by leveraging fine-grained semantic alignment between queries and video content.} 
\label{fig: framework}
\end{figure*}

\subsection{Granularity in Retrieval}
Fine-grained semantic granularity in both \textit{visual} and \textit{linguistic} streams is essential for accurate video grounding.
Early approaches relied on coarse frame-level cues or single-sentence queries~\cite{hendricks2017localizing, gao2017tall}, while later works introduced denser visual segments~\cite{zeng2020drn, song2021finegrained} and more elaborate query reasoning~\cite{yuan2019semantic}.
However, existing methods still align entire queries with generic captions or raw visual features, without explicitly managing semantic granularity.

Recent studies continue to exhibit this limitation. 
Diwan et al.~\cite{Diwan} improved ZVMR by 2.5× using off-the-shelf captions, yet these remain query-agnostic. 
Context-Enhanced VMR~\cite{liu2024contextvmr} incorporates BLIP-2 summaries, but lacks alignment with query intent. Moment-GPT~\cite{Moment-GPT} reformulates queries via LLaMA-3 and scores spans using frozen MLLMs, but still treats each query as a single undifferentiated unit and relies on query-agnostic captions, leading to potential semantic mismatches. 
Thus, recent methods fall short in achieving robust moment retrieval.
This persistent gap highlights the need for a framework that explicitly models and aligns content at complementary levels of granularity. Our work, GranAlign, is designed to address this very challenge.


\section{Method}
In this work, we propose a novel framework, GranAlign (illustrated in Fig.~\ref{fig: framework}), to establish a fine-grained semantic alignment between queries and video content, thereby enhancing the robustness and precision of zero-shot VMR.

\subsection{Overview}
Given an untrimmed video $V = \{v_i\}_{i=1}^{L_v}$ containing $L_v$ frames, and a textual query $Q = \{q_i\}_{i=1}^{L_q}$ comprising $L_q$ words, the goal of video moment retrieval (VMR) is to predict a set of temporal spans $T = \{t_s, t_e\} \in \mathbb{R}^{N_t \times 2}$. Here, $t_s$, $t_e$ denote the start and end times of a predicted segment, and $N_t$ is the number of predicted spans.

Our framework consists of three stages. First, to enable our Granularity-Aware Alignment, the input query is reformulated into simplified and detailed versions, and two corresponding caption sets--query-agnostic and query-aware--are generated (Sec.\ref{sec: scoring}). These are then used to compute a \textit{Granular Moment Score} for each video segment (Sec.\ref{sec: proposal}). Second, candidate temporal segments (\textit{moment proposals}) are generated based on these scores. Finally, a post-processing step including Non-Maximum Suppression (NMS) selects the most relevant segments.


\subsection{Granularity-Aware Alignment}
\label{sec: scoring}
This section introduces the core idea of our approach, \textit{Granularity-Aware Alignment}, which consists of three main components: granularity-based query rewriting, query-aware caption generation, and moment score computation.


\subsubsection{Granularity-based query rewriting}

\begin{figure*}[htb!]
\centerline{\includegraphics[width=0.95\textwidth]{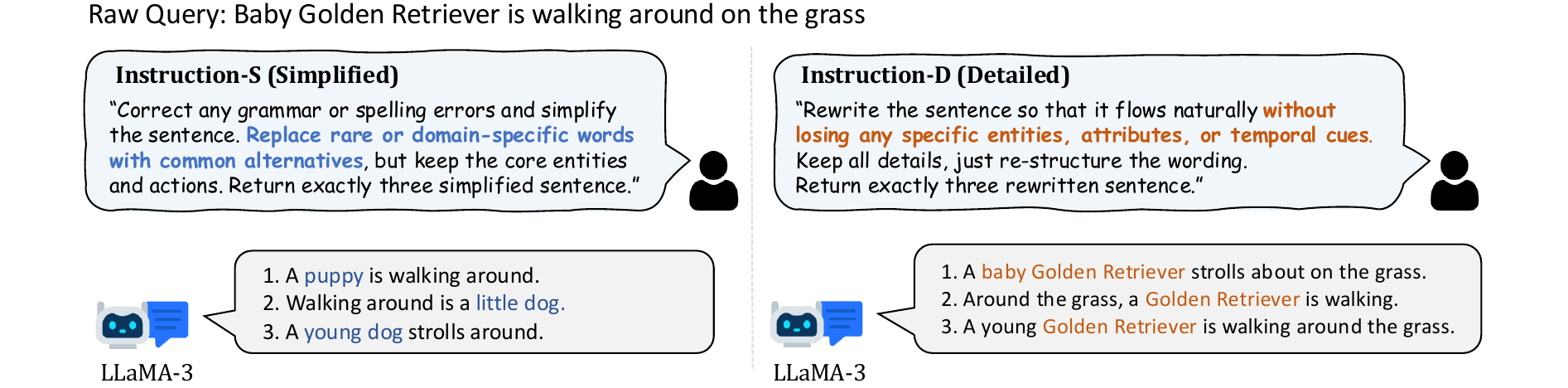}}
\caption{Examples of instruction pairs for granularity-based query rewriting. This figure illustrates sample prompts given to LLaMA-3 to generate simplified (recall-focused) and detailed (precision-focused) queries.}
\label{fig4}
\end{figure*}

To achieve more robust semantic alignment, we reformulate the input query $Q$ into semantically complementary representations via instruction-guided rewriting with LLaMA-3~\cite{llama3}.
Defining ``simple" and ``detail" with a single, rigid mathematical formula is challenging, as such definitions fail to capture the nuances of natural language. Therefore, rather than relying on a single prompt, we utilize multiple, manually crafted instruction pairs to generate the simplified query ($Q_s$) and the detailed query ($Q_d$). Figure~\ref{fig4} showcases representative examples from our instruction set.\footnote{The complete set of instruction pairs used in our experiments, along with a sensitivity analysis of their impact on performance, is available in Appendix E.} We empirically found that our framework's performance is robust and not overly sensitive to the specific instruction pair used.
%
%
The simplified query $Q_s$ is derived using a prompt that instructs the model to ``replace rare words with common alternatives" while ``keeping the core entities and actions."
To improve clarity, rare or incidental details are removed, and grammatical errors are corrected.
This form provides high generality and effectively retrieves broadly relevant candidate segments.
In contrast, the detailed query \( Q_d \) preserves fine-grained expressions, temporal context, and specific lexical choices.
These elements contribute to more precise alignment and accurate moment localization.

%

\subsubsection{Query-Aware Captioning}
Conventional video captioning methods generate descriptions from visuals alone, limiting their ability to accurately identify moments semantically aligned with a query.
Even a visually correct caption may lack the query intent needed for accurate retrieval.

To overcome this limitation, we aim to directly embed the query's intent into captions. However, applying this process to every frame is computationally prohibitive. We therefore propose a hybrid strategy that balances semantic precision with computational feasibility. This strategy first identifies a set of candidate frames by selecting the top-K\% with the highest similarity scores between the query $Q$ and each frame. Let $L_k$ denote the number of these candidate frames. Then, using Qwen2.5-VL~\cite{qwen2.5vl}, it generates two types of captions. First, a general, \textit{query-agnostic caption} ($C_{agn} \in \mathbb{R}^{L_v \times l}$) is created as a baseline for all $L_v$ frames. Second, a focused, \textit{query-aware caption} ($C_{awr} \in \mathbb{R}^{L_k \times l}$) is generated only for the $L_k$ candidate frames.

To create $C_{awr}$, we guide the generation process using key semantic elements extracted from the query. For instance, from ``A person picking up a pencil from the desk," we extract the entities \textit{\{person, pencil\}} and the action \textit{\{picking up\}} as semantic guidance. This hybrid strategy enables our framework to leverage the semantic precision of query-aware captions on the most critical regions while maintaining overall computational efficiency.

This process enables $C_{awr}$ to achieve stronger semantic alignment with the query compared to query-agnostic captions, which in turn improves the accuracy of moment localization. However, this approach can lead to failures, such as hallucinating visual content not present in the video or overly mimicking the query's linguistic structure. 
Therefore, our final moment scoring process is designed to leverage the advantages of $C_{awr}$ while mitigating these potential failures.


\subsubsection{Granular Moment scoring}

\begin{figure*}[htb!]
    \centerline{\includegraphics[width=0.95\textwidth]{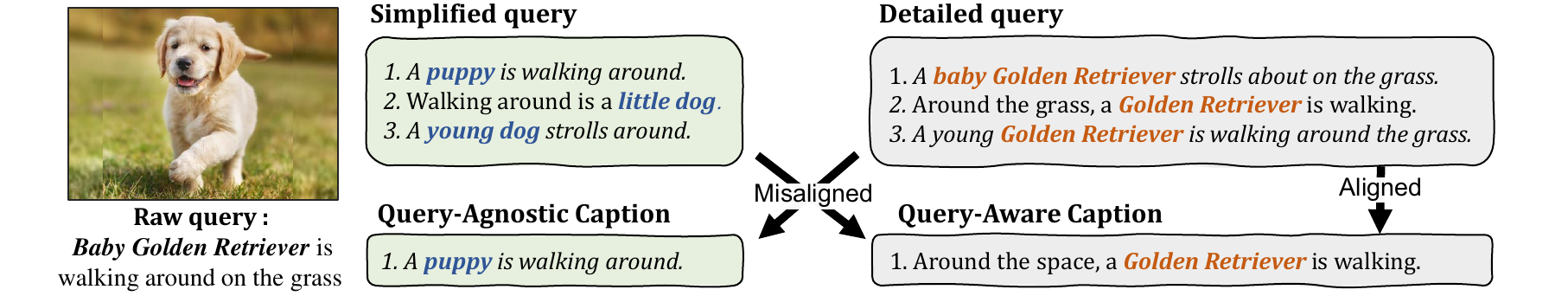}}
    \caption{Illustration of the semantic alignment patterns at different granularity levels. A simplified query (e.g., ``a puppy'') aligns well with a query-agnostic caption, whereas a detailed query (e.g., ``Golden Retriever'') requires a query-aware caption. Mismatches in granularity can lead to failed alignments.}
    \label{fig5}
\end{figure*}

The two query-caption pairs, simplified-agnostic $(Q_s, C_{\text{agn}})$ and detailed-aware $(Q_d, C_{\text{awr}})$, exhibit distinct characteristics with complementary strengths and limitations, as illustrated in Fig~\ref{fig5}. The simplified query $Q_s$ and generic caption $C_{\text{agn}}$ offer broad coverage of semantically similar scenes, yielding higher recall but often lacking the precision to localize the exact moment. In contrast, the detailed query $Q_d$ and query-aware caption $C_{\text{awr}}$ enable more accurate alignment via fine-grained semantics, but are more prone to errors like hallucination or misalignment from over-relying on the query. This highlights the importance of matching granularity levels for successful alignment, which is the core principle of our approach.


To leverage the complementary benefits of both pairs, we compute a composite \textit{moment score} by integrating the semantic similarities of ($Q_s$, $C_{{agn}}$) and ($Q_d$, $C_{{awr}})$. Specifically, we evaluate each candidate moment's semantic alignment using both query-caption pairs and take the average as the final score. The combined frame-level similarity score $S_f$ is defined as:



\begin{equation}
S_f = \frac{1}{2m} \sum_{i=1}^{m} \left[
\textstyle \operatorname{g}(q_{s}^{(i)}, C_{agn,f}) +
\operatorname{g}(q_{d}^{(i)}, C_{awr,f})
\right]
\label{eq:frame_score}
\end{equation}


The similarity score $S_f$ is computed for each frame $f$ in the video by averaging the semantic similarity between rephrased queries and their corresponding captions.
Each rephrased query pair consists of a simplified query $q_{s}^{(i)}$ and a detailed query $q_{d}^{(i)}$, both derived from the original input query through controlled rewriting, where $m$ denotes the total number of such pairs.
$C_{agn,f}$ and $C_{awr,f}$ represent the caption embeddings for frame $f$ in the query-agnostic and query-aware caption sets, respectively.
Here, $\operatorname{g}(\cdot, \cdot)$ denotes the normalized cosine similarity between the sentence embeddings of the query and the caption.

This formulation mitigates potential biases or false positives that may arise from relying on a single query-caption pair, enabling more robust and reliable moment retrieval. We compute the similarity by first embedding queries and captions with a SentenceTransformer model~\cite{sentencebert}, and then calculating their cosine similarity. The resulting frame-level score $S_f$ is then used to generate and rank candidate segments.
 

\subsection{Moment Proposal Generation}
\label{sec: proposal}
In this stage, the computed \textit{Moment Scores} are used to generate candidate segments via a \textit{Moment Proposal Generator}, followed by a \textit{Post Processing} step to refine the final output.


\subsubsection{Moment proposal generator}
The Moment Proposal Generator (MPG) creates semantically coherent candidate spans from frame-level similarity scores $S_f$. To prevent a single continuous event from being fragmented into multiple short clips, MPG merges adjacent high-scoring frames into a single proposal if the temporal gap between them is within a threshold of $\tau$ frames. While merging frames based on the $\tau$ ensures continuity, it can create diluted proposals where high-scoring keyframes are connected by a series of low-relevance frames. To mitigate this issue, spans with an average similarity in the bottom n\% are discarded to maintain relevance. This acts as a crucial quality control step, ensuring that only proposals with a consistently high level of semantic relevance proceed to the next stage.

We score each candidate span $p$ based on its semantic relevance and a length regularization. The first term, $\mu_p$, is the average semantic similarity of all frames within the span. The second term, $\rho_p$, is the length of span $p$ normalized by the total length of all candidate spans. This normalized length $\rho_p$ serves as a crucial regularization factor to penalize overly long proposals that might have low average relevance and to prevent a bias towards trivial, short segments, ensuring that the final candidates are of a meaningful duration.
The final score for each candidate span $p$ is computed as:
\begin{equation}
\text{Score}(p) = (1 - \lambda)\mu_p + \lambda\rho_p,
\quad 0 \le \lambda \le 1.
\label{eq:span_score}
\end{equation}
The parameter $\lambda$ (set to 0.3) balances this trade-off. This regularized scoring provides more robust candidates than simpler threshold- or length-based approaches.


\subsubsection{Post processing}
All candidate spans generated by the \textit{Moment Proposal Generator} are collected into a set, with each span defined by its start and end frame indices.
Any two spans $p_i$ and $p_j$ in this set represent distinct candidate intervals.
To refine the set of candidate spans, we apply \textit{Non-Maximum Suppression} (NMS), which iteratively discards the lower-scoring of any two spans whose Intersection-over-Union (IoU) exceeds a threshold $\theta_{\mathrm{NMS}}$, yielding the final moment prediction $\{t_s, t_e\}$.


\section{Experiments}
This section details the experiments conducted to validate the effectiveness of the proposed \textbf{GranAlign} framework. We present the experimental setup, a performance comparison against state-of-the-art methods, an ablation study analyzing the contributions of our core components, and a further analysis to demonstrate the robustness of our method.

\noindent\textbf{Datasets}
We evaluate on public benchmarks: \textit{QVHighlights}~\cite{QVH}, \textit{Charades-STA}~\cite{Charades}, and \textit{ActivityNet-Captions}~\cite{ActivityNet-Captions}.

\noindent\textbf{Evaluation Metrics}
Following prior work, we report standard VMR scores, including Recall@1 (R1@$\text{n}$), the percentage of queries where the top prediction is correct at a given IoU threshold n, and mean Average Precision (mAP@$\text{n}$), which evaluates the overall localization precision. Full statistics for the datasets and detailed definitions of the metrics are provided in Appendix D.


\subsection{Comparison with the State-of-the-Art}
Table~\ref{tab1} shows that GranAlign consistently surpasses the previous zero-shot SOTA \cite{Moment-GPT} on QVHighlights.
On the validation split, it improves all metrics by \textbf{+3.04\% to +3.93\%}, with the largest gain in mAP@0.5; on the hidden test set, it still leads by \textbf{+1.6\% to +3.84\%}.
These gains confirm that granularity-based query rewriting and query-aware captioning enhance semantic alignment and, consequently, retrieval accuracy.
Table~\ref{tab2} shows strong transfer to Charades-STA and ActivityNet-Captions.
On Charades-STA, GranAlign exceeds the previous SOTA by +1.2\% (R1@0.5) and +1.5\% (mIoU); on ActivityNet-Captions, it gains +2.9\% (R1@0.5) and +2.3\% (mIoU).
Thus, the proposed alignment strategy scales from short to long, diverse videos, providing robust zero-shot moment retrieval across benchmarks.


\begin{table*}[t]
    \centering
    \resizebox{1\textwidth}{!}{%
    \begin{tabular}{lccccccccccc}
    \toprule
    \textbf{Method} & \textbf{MLLM} & \textbf{Setting}
        & \multicolumn{4}{c}{\textbf{QVHighlights test}} & \multicolumn{4}{c}{\textbf{QVHighlights val}} \\
    \cmidrule(r){4-7} \cmidrule(r){8-11}
     & &
        & \textbf{R1@0.5} & \textbf{R1@0.7} & \textbf{mAP@0.5} & \textbf{mAP@avg}
        & \textbf{R1@0.5} & \textbf{R1@0.7} & \textbf{mAP@0.5} & \textbf{mAP@avg} \\
    \midrule
    VTimeLLM~\cite{vtimellm}      & \checkmark & FS
        & 47.2 & 29.3 & 47.3 & 27.4
        & 48.8 & 29.5 & 49.3 & 26.8 \\
    LLaViLo~\cite{LLaViLo}         & \checkmark & FS
        & 48.6 & 29.7 & 48.7 & 27.9
        & 49.0 & 30.4 & 49.4 & 28.9 \\
    Moment-DETR~\cite{QVH}         & --          & FS
        & 52.9 & 33.0 & 54.8 & 30.7
        & 54.2 & 33.4 & 55.4 & 31.1 \\
    MomentDiff~\cite{momentdiff}         & --          & FS
        & - & - & - & -
        & 57.8 & 39.2 & 54.6 & 35.3 \\        
    \midrule
    CPL~\cite{CPL}                & --          & WS
        & 30.8 & 10.8 & 22.8 & --
        & --   & --   & --   & --   \\
    CPI~\cite{kong2023dynamic}    & --          & WS
        & 32.3 & 11.8 & 23.7 & --
        & --   & --   & --   & --   \\
    \midrule
    Liu et al.~\cite{Liu}         & --          & US
        & --   & --   & --   & --
        & 12.3 & 3.5  & 10.4 & 2.7  \\
    PZVMR~\cite{PZVMR}           & --          & US
        & 14.2 & 4.9  & 15.7 & 4.6
        & 12.6 & 5.1  & 16.2 & 5.3  \\
    \midrule
    VideoLLaMa~\cite{5_VideoLLaMA}         & \checkmark & ZS
        & 17.1 & 6.7  & 18.2 & 6.2
        & 18.5   & 6.9   & 17.8   & 7.1   \\
    VideoChatGPT~\cite{VideoChatGPT}         & \checkmark & ZS
        & 21.1 & 10.2  & 22.8 & 9.5
        & 22.4   & 10.8   & 21.9   & 10.3   \\        
    UniVTG~\cite{UniVTG}         & \  & ZS
        & 25.2 & 9.0  & 27.4 & 10.9
        & --   & --   & --   & --   \\
    Diwan~\cite{Diwan}           & \checkmark & ZS
        & --   & --   & --   & --
        & 48.3 & 31.0 & 47.3 & 28.0 \\
    Moment-GPT~\cite{Moment-GPT} & \checkmark & ZS
        & 58.3 & 37.7 & 55.1 & 35.0
        & 58.9 & 38.6 & 55.7 & 35.9 \\
    GranAlign (ours)                  & \checkmark & ZS
        & \textbf{59.92} & \textbf{39.3} & \textbf{58.94} & \textbf{38.23}
        & \textbf{61.94} & \textbf{41.81} & \textbf{59.63} & \textbf{39.12} \\
    \bottomrule
    \end{tabular}%
    }
    \caption{Performance comparison on QVHighlights test and val sets. ``MLLM" indicates the use of a multimodal large language model, and ``FS/WS/US/ZS" denote fully-supervised, weakly-supervised, unsupervised, and zero-shot settings. }
    \label{tab1}
\end{table*}

\begin{table*}[t]
    \centering
    \resizebox{1\textwidth}{!}{%
    \begin{tabular}{lccccccccccc}
    \toprule
    \textbf{Method} & \textbf{MLLM} & \textbf{Setting}
        & \multicolumn{4}{c}{\textbf{Charades-STA}} & \multicolumn{4}{c}{\textbf{ActivityNet-Captions}} \\
    \cmidrule(r){4-7} \cmidrule(r){8-11}
     & & 
        & \textbf{R1@0.3} & \textbf{R1@0.5} & \textbf{R1@0.7} & \textbf{mIoU}
        & \textbf{R1@0.3} & \textbf{R1@0.5} & \textbf{R1@0.7} & \textbf{mIoU} \\
    \midrule
    GroundingGPT~\cite{groundinggpt}      & \checkmark & FS
        & --   & 29.6 & 11.9 & --
        & --   & --   & --   & --   \\
    VTimeLLM~\cite{vtimellm}              & \checkmark & FS
        & 55.3 & 34.3 & 14.7 & 34.6
        & 44.8 & 29.5 & 14.2 & 31.4 \\
    TimeChat~\cite{timechat}              & \checkmark & FS
        & --   & 43.8 & 22.7 & --
        & --   & --   & --   & --   \\
    Moment-DETR~\cite{QVH}                & --          & FS
        & 62.1 & 48.2 & 25.3 & 42.3
        & 52.6 & 32.5 & 15.3 & 37.8 \\
    \midrule
    CPL~\cite{CPL}                        & --          & WS
        & 56.0 & 38.1 & 20.3 & 37.8
        & 52.4 & 30.9 & 12.0 & 32.6 \\
    Huang et al.~\cite{huang2023}        & --          & WS
        & 59.2 & 44.2 & 22.1 & 39.4
        & 54.8 & 32.9 & --   & 36.4 \\
    \midrule
    PSVL~\cite{nam2021}                   & --          & US
        & 45.2 & 30.9 & 14.2 & 30.9
        & 45.1 & 29.8 & 15.73 & 30.2 \\
    Liu et al.~\cite{Liu}             & --          & US
        & 44.2 & 28.7 & 14.7 & --
        & 47.3 & 28.2 & --   & --   \\
    \midrule
    TimeChat~\cite{timechat}              & \checkmark & ZS
        & --   & 32.2 & 13.4 & --
        & --   & --   & --   & --   \\
    Luo et al.~\cite{luo2024}            & \checkmark & ZS
        & 53.4 & 36.0 & 19.3 & 34.1
        & 45.6 & 27.4 & 12.3 & 28.4 \\
    Moment-GPT ~\cite{Moment-GPT}   & \checkmark & ZS
        & 58.2 & 38.4 & 21.6 & 36.5
        & 48.1 & 31.1 & 14.9 & 30.8 \\
    GranAlign (Ours)   & \checkmark & ZS
        & \textbf{59.1} & \textbf{39.6} & \textbf{22.7} & \textbf{38.0}
        & \textbf{50.3} & \textbf{34.0} & \textbf{16.5} & \textbf{33.1} \\        
    \bottomrule
    \end{tabular}%
    }
    \caption{Performance comparison on Charades-STA and ActivityNet-Captions under various supervision settings.}
    \label{tab2}
\end{table*}


\begin{table}[h]
  \centering
  \resizebox{0.85\columnwidth}{!}{%
  \begin{tabular}{c c c c c c c}
    \toprule
    \textbf{$C_{{agn}}$} & \textbf{$C_{{awr}}$} & \textbf{R1@0.5} & \textbf{R1@0.7} & \textbf{mAP@0.5} & \textbf{mAP@avg} \\
    \midrule
    $Q_r$ &   -   & 57.94          & 38.84          & 51.64          & 31.8             \\
       -  & $Q_r$ & 58.19          & 39.03          & 52.12          & 32.13              \\
    \midrule
    $Q_s$ &   -   & 58.97          & 40.71          & 56.88          & 37.13          \\
       -  & $Q_s$ & 57.55          & 38.39          & 56.02          & 36.22             \\
    $Q_d$ &   -   & 58.19          & 39.42          & 56.4              & 36.82              \\
       -  & $Q_d$ & 59.48          & 40.52          & 57.01              & 37.65      \\
    \midrule
    $Q_s$ & $Q_s$ & 61.03          & \textbf{42.06}              & 59.12          & 38.78       \\
    $Q_d$ & $Q_s$ & 60.90          &  41.26             & 58.83          & 38.63          \\
    $Q_s$ & $Q_d$ & \textbf{61.94} & 41.81              & \textbf{59.63} & \textbf{39.12} \\
    $Q_d$ & $Q_d$ & 61.19          & 41.35              & 59.52          & 39.01          \\
    \bottomrule
  \end{tabular}%
  }
  \caption{Consolidated ablation study results. We denote queries as $Q$ and captions as $C$, with subscripts for raw ($r$), simplified ($s$), detailed ($d$), query-agnostic ($agn$), and query-aware ($awr$). The ($Q_s$, $Q_d$) row represents our full model.}
  \label{tab:final_ablation}
\end{table}

\subsection{Ablation Studies}
We perform an ablation study on the QVHighlights validation set to evaluate the contribution of each component in our framework, with all results consolidated in Table~\ref{tab:final_ablation}.

\subsubsection{Notation} In the following ablation experiments, we revisit the notation for clarity: $Q_r$ denotes the raw query, $Q_s$, the simplified query; $Q_d$, the detailed query; $C_{agn}$, the query-agnostic caption; and $C_{awr}$, the query-aware caption.


\subsubsection{Impact of Query-Aware Caption}
As shown in Table~\ref{tab:final_ablation}, using a query-aware caption ($C_{awr}$) with the raw query yields consistent improvements over a query-agnostic one ($C_{agn}$) across all metrics. This suggests that tailoring captions to the query content enhances semantic alignment.


\subsubsection{Effectiveness of Granularity-Aligned Pairing}
We analyze the effects of granularity alignment in Table~\ref{tab:final_ablation}. Performance declines when query and caption granularities are mismatched (e.g., pairing $Q_s$ with $C_{awr}$), whereas aligned pairs (e.g., pairing $Q_d$ with $C_{awr}$) perform better.

However, these single aligned pairs present a trade-off: the simplified pair ($Q_s$, $C_{agn}$) offers high recall but low precision, while the detailed pair ($Q_d$, $C_{awr}$), despite achieving the best performance among single pairs, provides high precision but can fail on subtle mismatches.
Our GranAlign framework resolves this by combining both pairs, leveraging their complementary strengths. As a result, GranAlign achieves the best overall performance with consistent gains across all metrics. 
We also confirm this approach's robustness with a BLIP-2 backbone (please see Appendix A).




\subsubsection{Implementation Details}
We implementd our framework using LLaMA3-8B~\cite{llama3} for query rewriting and Qwen2.5-VL-7B~\cite{qwen2.5vl} for caption generation, with initial frames filtered by CLIP ViT-B/32~\cite{CLIP}. All experiments were run on four NVIDIA A6000 GPUs, and a comprehensive list of implementation details is available in Appendix B.

\begin{figure}[htb!]  \centerline{\includegraphics[width=1\columnwidth]{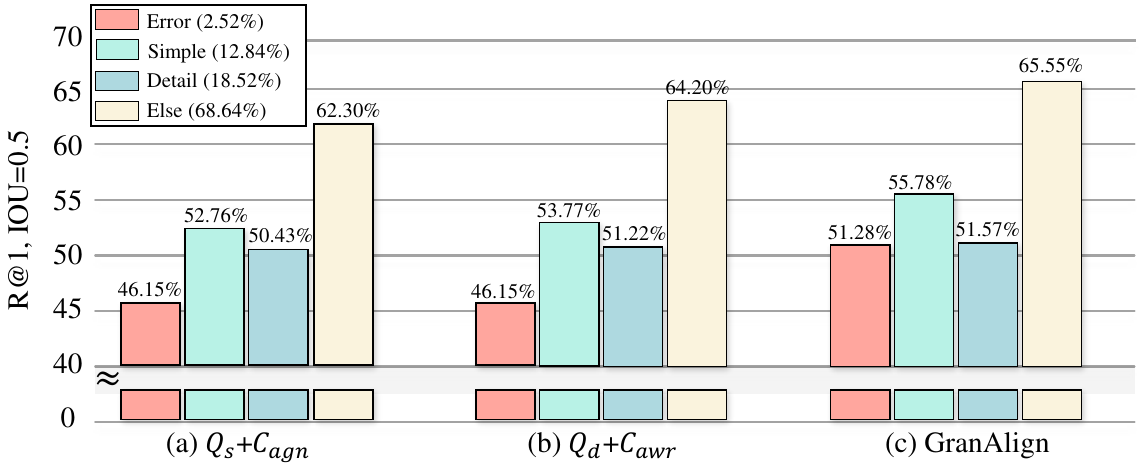}}
    \caption{Analysis of performance across simplified, detailed, error, and other query types on QVHighlights val. The Y-axis is truncated below 40 for clarity. }
    \label{fig:analysis}
\end{figure}
\begin{figure}[htb!]
\centerline{\includegraphics[width=1\columnwidth]{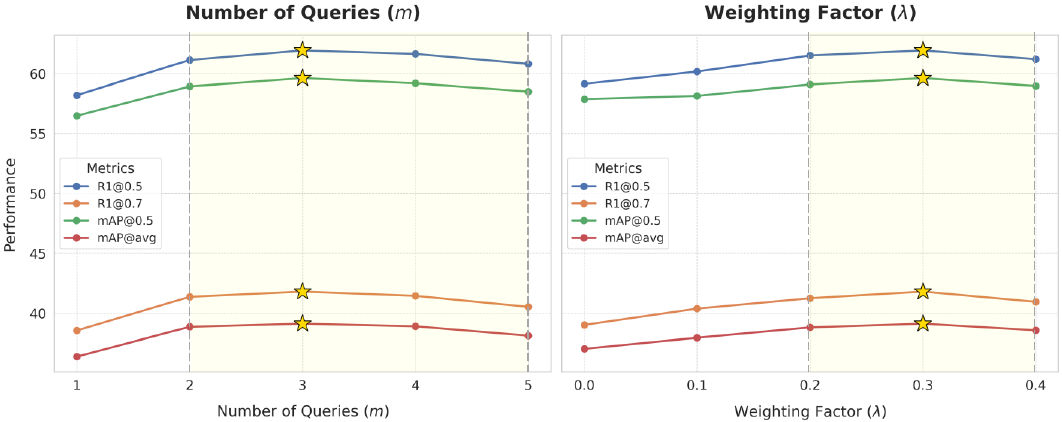}}
    \caption{Analysis of performance across variant hyperparameters. The yellow shaded regions highlight the parameter range where our framework demonstrates robust and stable performance around the optimal settings (marked by stars).}
    \label{fig:sensitive}
\end{figure}

\subsection{Further Analysis}
\subsubsection{Robustness to Query Variation} 
As shown in Fig~\ref{fig:analysis}, we evaluated GranAlign's robustness across four linguistically diverse query types (Error, Simplified, Detailed, and Else).\footnote{Queries were classified by length: Simple ($\le$6 tokens), Detail ($\ge$20 tokens or containing a proper noun), or Else (all others). Independently, queries with grammatical errors were flagged as Error.}
The baseline combinations, (a) ($Q_s$, $C_{{agn}}$) and (b) ($Q_d$, $C_{{awr}}$), exhibit complementary but limited behaviors. The former provides broad recall but lacks precision, while the latter captures fine-grained semantics yet suffers from overfitting to the query, leading to hallucinations or misalignment. 
In contrast, our proposed GranAlign integrates both simplified and detailed query reformulations with query-aware captioning, achieving robust performance across all categories by effectively balancing recall and precision.
%


\subsubsection{Sensitivity Analysis}
As shown in Figure~\ref{fig:sensitive}, our novel framework, GranAlign, is robust to variant hyperparameter settings. For example, performance peaks when using 3 rewritten queries (m=3) and a weighting factor of 0.3 ($\lambda$=0.3), while remaining stable across a range of nearby values. Our full experiments on hyperparameter selection are detailed in Appendix C.

\subsubsection{Efficiency Analysis}
In Table~\ref{tab:my_label}, we compare the training cost, GPU memory usage, and inference time against fully supervised and zero-shot methods. As a zero-shot method, GranAlign incurs no training cost and, for efficiency, pre-generates query-agnostic captions offline.
\begin{table}[h]
  \centering
  \resizebox{\columnwidth}{!}{%
  \begin{tabular}{@{}lcccccc@{}}
    \toprule
    Setup & Methods & R1@0.5 & mAP & TrC & InT (s) & GMU (G) \\
    \midrule
    ZS & VideoChatGPT~\cite{VideoChatGPT} & 22.4 & 10.3 & 0 & 9.8 & 11 \\
    FS & VTimeLLM~\cite{vtimellm} & 48.8 & 26.8 & 4090 40h & 11.2 & 18 \\
    ZS & Wattersseril~\cite{wattasseril} & 53.1 & 30.2 & 0 & 12.7 & 14 \\
    ZS & Moment-GPT~\cite{Moment-GPT} & 58.9 & 35.9 & 0 & 16.1 & 32 \\
    ZS & Ours & \textbf{61.94} & \textbf{39.12} & \textbf{0} & \textbf{6.2} & \textbf{22} \\
    \bottomrule
  \end{tabular}}
  \caption{Comparison of training cost (TrC), GPU memory usage (GMU), and inference time (InT) on the QVHighlights dataset.}
  \label{tab:my_label}
\end{table}
\begin{figure}[h]
    \centerline{\includegraphics[width=1\columnwidth]{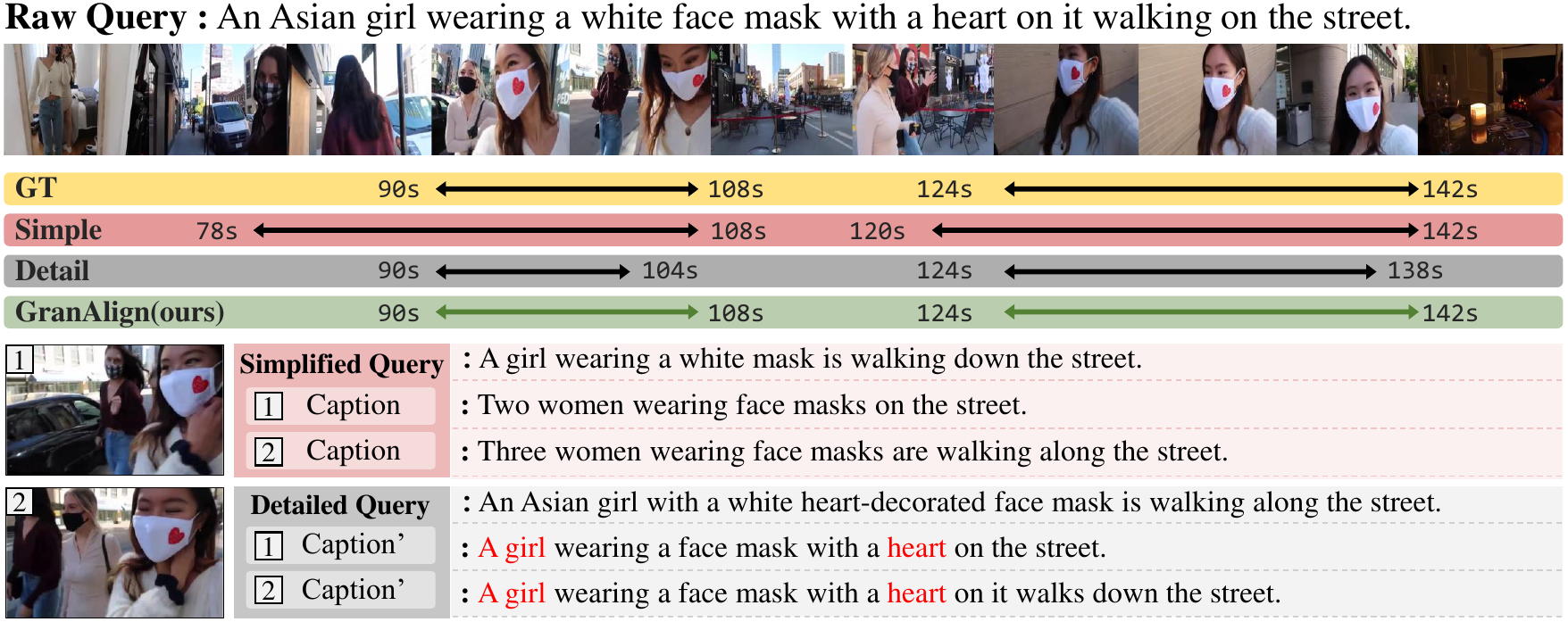}}
    \caption{Qualitative results on QVHighlights. Caption' in Detailed Query denotes query-aware caption.}
    \label{fig8}
\end{figure}
Subsequently, during the online process, it selectively generates additional captions only for key frames that exceed a certain threshold. This two-stage approach contributes to reducing inference time and helps the model maintain a competitive memory footprint. 
As a result, GranAlign achieves the best performance, outperforming other configurations on most key metrics such as R1@0.5 and mAP while also recording the shortest inference time.

\subsubsection{Qualitative Results}
Fig~\ref{fig8} qualitatively illustrates how GranAlign leverages semantic granularity to accurately localize moments in a zero-shot setting.
Given the query ``An Asian girl wearing a white face mask with a heart on it walking on the street," the ground truth (GT) segment spans from 90 to 142 seconds.
The Simple path, which aligns a simplified query with query-agnostic captions, fails to capture specific attributes and instead relies on general information, resulting in an over-extended prediction starting from 78s, thus including irrelevant content.
In contrast, the Detail path, which aligns a detailed query with a query-aware caption, identifies fine-grained cues such as ``heart" and begins at a more precise point, but slightly under-segments the end (missing frames after 138s).
GranAlign, by integrating the alignment scores from both paths, successfully predicts the exact GT segment (90-142s), achieving the best of both approaches.
This qualitative result provides clear evidence that our granular design is key to overcoming the core trade-off between coverage and precision, leading to more accurate and robust moment localization.


\section{Discussions and Limitations}
While GranAlign achieves strong performance, its limitations stem from the generative nature of its core components, presenting clear avenues for future research. For instance, the risk of hallucination in query-aware captioning could be mitigated by a fact-checking mechanism that verifies generated details against the visual evidence. Similarly, a semantic verification step could ensure that LLM-based query rewriting preserves the original intent. Pursuing these enhancements would further improve the reliability and generalizability of granularity-aware moment retrieval systems.

\section{Conclusion}

We proposed Granularity-Aware Alignment (GranAlign), a novel, training-free framework for ZVMR. By aligning queries and video content at multiple levels of semantic granularity, GranAlign resolves the trade-off between coverage and precision, balancing the high recall of general representations with the high precision of detailed attributes. As a result, GranAlign establishes a SOTA across major benchmarks without task-specific training cost, presenting new possibilities for zero-shot VMR. We believe our granularity-aware alignment strategy offers a promising and effective direction for future multimodal understanding systems.


\appendix

\section*{Appendix}
\label{sec:appendix}

\section{Further Analysis}
\subsection{Generalization to Downstream Tasks}


To further validate the effectiveness of our granularity-aware design, we conduct an ablation on the Video Highlight Detection (VHD) task, as shown in Table~\ref{tab:vhd_performance}. We first evaluate GranAlign\dag, a variant that uses only the detailed query with query-aware captioning ($Q_d$, $C_{{awr}}$). While this setting already achieves strong performance, the full GranAlign model-combining both simplified-original ($Q_s$, $C_{{agn}}$) and detailed-aware ($Q_d$, $C_{{awr}}$) pairings-further improves the results, reaching 39.35\% mAP and 66.34\% HIT@1. Notably, GranAlign outperforms even the fully supervised QD-DETR (mAP 39.04\%, HIT@1 62.87\%), demonstrating that leveraging complementary granularity pairs contributes significantly to retrieval accuracy. These results confirm that our dual-granularity alignment strategy is not only essential for moment retrieval but also broadly transferable to related downstream tasks.

\subsection{Granularity-Aware Pairing}
To verify that the same granularity patterns from our ablation study persist when employing BLIP2-generated captions, we rerun the ablation with BLIP2 captions. As shown in Table~\ref{tab:blip2_ablation}, the mixed‐granularity settings (b) and (c) again outperform the homogeneous ones, with configuration (c) achieving the highest average mAP, thereby confirming GranAlign's robustness across different captioning backbones.

\section{Implementation Details}

This section provides a detailed description of the implementation details and hyperparameters used in our method.

\noindent (i) Frame-level captions are generated using Qwen2.5-VL-7B~\cite{qwen2.5vl} at a sampling rate of 0.5 fps.
We then extract features from both frames and the query using the CLIP ViT-B/32 model\cite{CLIP} and select only those captions whose frame-level similarity ranks within the top 10\% (K) for further processing.

\begin{table}[h]
  \centering
  \resizebox{1\columnwidth}{!}{%
  \small
  \begin{tabular}{lccc}
    \toprule
    \textbf{Model}                         & \textbf{Setting} & \textbf{mAP} & \textbf{HIT@1} \\
    \midrule
    TimeChat~\cite{timechat}       & FS               & 14.5         & 23.9           \\
    Moment-DETR~\cite{QVH}         & FS               & 35.7         & 55.7           \\
    QD-DETR~\cite{35_QD-DETR}      & FS               & 39.04        & 62.87          \\
    \midrule
    Moment-GPT~\cite{Moment-GPT}                     & ZS               & 36.7        & 62.7          \\    
    GranAlign\dag                     & ZS               & 37.41        & 64.98          \\
    GranAlign (ours)               & ZS               & ~\textbf{39.35} & ~\textbf{66.34}           \\        
    \bottomrule
  \end{tabular}%
  }
  \caption{VHD performance comparison under fully-/zero-shot settings.}
  \label{tab:vhd_performance}
\end{table}
\begin{table}[h]
  \centering
  \renewcommand{\arraystretch}{1.1}
  \resizebox{\columnwidth}{!}{%
  \begin{tabular}{c c c c c c c c c}
    \toprule
    \textbf{Config.} & \textbf{$C_{{agn}}$} & \textbf{$C_{{awr}}$}
      & \textbf{mIoU} & \textbf{R1@0.5} & \textbf{R1@0.7}
      & \textbf{mAP@0.5} & \textbf{mAP@0.75} & \textbf{mAP@avg} \\
    \midrule
    (a) & $Q_s$ & $Q_s$ & 54.53 & 58.00 & 38.71 & 55.05 & 36.33 & 36.20 \\
    (b) & $Q_d$ & $Q_s$ & 54.50 & 57.74 & 38.45 & 55.16 & 36.36 & 36.30 \\
    (c) & $Q_s$ & $Q_d$ & 56.00 & 59.32 & 40.68 & 56.71 & 38.05 & 37.80 \\
    (d) & $Q_d$ & $Q_d$ & 55.05 & 58.45 & 39.81 & 55.71 & 37.01 & 36.73 \\
    \bottomrule
  \end{tabular}%
  }
  \caption{BLIP2 Ablation results under different \texttt{$C_{{agn}}$}/\texttt{$C_{{awr}}$} settings.}
  \label{tab:blip2_ablation}
\end{table}

\noindent (ii) Query rewriting is performed using LLaMA3-8B~\cite{llama3}. 
Empirically, we found that generating 3 rewritten queries ($m$) yields the best performance.  
The rewritten query $Q'$ is designed to preserve the semantic meaning of the original query $Q$.

\noindent (iii) For candidate span scoring, we use an inverse cumulative histogram based on the similarity distribution.
The distribution is divided into 10 bins, and candidates within the top 8 are selected.
Span selection is dynamically adjusted to include borderline cases based on the score distribution.
In addition, gaps of up to  $\tau$=6 consecutive frames with low similarity between candidate spans are merged into a single span.
Here, similarity is computed between the debised query and both conditionally-processed and original queries, and spans are excluded from merging if their similarity falls within the bottom 20\% (n) of the distribution.

\noindent (iv) The initial and second-stage span selection thresholds are empirically determined based on validation performance.
The weighting factor $\lambda$ in the Span Scorer is set to 0.3.

\noindent (v) IoU-based Non-Maximum Suppression (NMS) is applied for final span selection with a threshold of 0.9 ($\theta_{NMS}$).
All experiments are conducted using 4 NVIDIA A6000 48GB GPUs.

\section{Hyperparameter Analysis}
This section presents the ablation studies conducted to determine the optimal values for the hyperparameters in our framework.

\begin{table}[h!]
\centering
\resizebox{0.8\columnwidth}{!}{%
\label{tab:ablation_queries}
\begin{tabular}{lcccc}
\toprule
$m$ & R1@0.5 & R1@0.7 & mAP@0.5 & mAP@avg \\
\midrule
1 & 58.21 & 38.55 & 56.49 & 36.38 \\
2 & 61.14 & 41.37 & 58.94 & 38.86 \\
\textbf{3}& \textbf{61.94} & \textbf{41.81} & \textbf{59.63} & \textbf{39.12} \\
4 & 61.65 & 41.45 & 59.20 & 38.90 \\
5 & 60.82 & 40.53 & 58.51 & 38.12 \\
\bottomrule
\end{tabular}}
\caption{Ablation on the number of queries ($m$) on the QVHighlights validation set.}
\end{table}

\begin{table}[h!]
\centering
\resizebox{0.8\columnwidth}{!}{%
\label{tab:ablation_lambda}
\begin{tabular}{lcccc}
\toprule
$\lambda$ & R1@0.5 & R1@0.7 & mAP@0.5 & mAP@avg \\
\midrule
0.0 & 59.15 & 39.02 & 57.88 & 37.03 \\
0.1 & 60.18 & 40.38 & 58.15 & 37.95 \\
0.2 & 61.52 & 41.25 & 59.10 & 38.81 \\
\textbf{0.3} & \textbf{61.94} & \textbf{41.81} & \textbf{59.63} & \textbf{39.12} \\
0.4 & 61.21 & 40.95 & 58.98 & 38.57 \\
\bottomrule
\end{tabular}}
\caption{Ablation on the weighting factor ($\lambda$) on the QVHighlights validation set.}
\end{table}

\begin{table}[h!]
\centering
\resizebox{0.65\columnwidth}{!}{%
\label{tab:ablation_caption_ratio}
\begin{tabular}{lccc}
\toprule
K (\%) & R1@0.3 & R1@0.5 & mIoU \\
\midrule
5\%  & 58.2 & 38.5 & 37.1 \\
\textbf{10\%} & \textbf{59.1} & \textbf{39.6} & \textbf{38.0} \\
20\% & 58.8 & 39.1 & 37.7 \\
30\% & 57.9 & 38.2 & 36.9 \\
\bottomrule
\end{tabular}}
\caption{Ablation on the caption selection ratio (Top-K\%) on the Charades-STA dataset.}
\end{table}

\begin{table}[h!]
\centering
\resizebox{0.6\columnwidth}{!}{%
\label{tab:ablation_merging_gap}
\begin{tabular}{lccc}
\toprule
$\tau$ & R1@0.3 & R1@0.5 & mIoU \\
\midrule
0  & 57.5 & 37.9 & 36.5 \\
2  & 58.6 & 39.0 & 37.4 \\
4  & 58.9 & 39.4 & 37.8 \\
\textbf{6}  & \textbf{59.1} & \textbf{39.6} & \textbf{38.0} \\
8 & 58.3 & 38.8 & 37.2 \\
\bottomrule
\end{tabular}}
\caption{Ablation on the span merging gap ($\tau$) on the Charades-STA dataset.}
\end{table}

\begin{table}[h!]
\centering
\resizebox{0.7\columnwidth}{!}{%
\label{tab:ablation_exclusion_threshold}
\begin{tabular}{lccc}
\toprule
n (\%) & R1@0.3 & R1@0.5 & mIoU \\
\midrule
Bottom 10\% & 58.7 & 39.2 & 37.6 \\
\textbf{Bottom 20\%} & \textbf{59.1} & \textbf{39.6} & \textbf{38.0} \\
Bottom 30\% & 58.2 & 38.6 & 37.1 \\
Bottom 40\% & 57.6 & 38.0 & 36.4 \\
\bottomrule
\end{tabular}}
\caption{Ablation on the merging exclusion threshold on the Charades-STA dataset.}
\end{table}

\begin{table}[h!]
\centering
\resizebox{0.65\columnwidth}{!}{%
\label{tab:ablation_nms}
\begin{tabular}{lccc}
\toprule
$\theta_{NMS}$ & R1@0.3 & R1@0.5 & mIoU \\
\midrule
0.8 & 58.1 & 38.8 & 37.0 \\
0.85 & 58.8 & 39.3 & 37.7 \\
\textbf{0.9} & \textbf{59.1} & \textbf{39.6} & \textbf{38.0} \\
0.95 & 58.9 & 39.5 & 37.9 \\
\bottomrule
\end{tabular}}
\caption{Ablation on the NMS IoU threshold ($\theta_{NMS}$) on the Charades-STA dataset.}
\end{table}



\section{Datasets, Evaluation Metrics}
\subsubsection{Datasets}
\label{sec: data}

QVHighlights \cite{QVH} is a benchmark dataset of 10,148 YouTube videos, each about 150s long, with 7,218/1,550/1,542 train/val/test queries. Each query has one or more annotated moments. Test labels are hidden; evaluation is via a public server.

Charades-STA \cite{Charades} contains 9,848 indoor videos, averaging 31 s, and 16,128 query-moment pairs. We follow the standard 12,408/3,720 train/test split and sample 10\% of the training data for validation.

ActivityNet-Captions \cite{ActivityNet-Captions} is a benchmark of 20,000 untrimmed videos totaling approximately 849 hours, paired with about 100,000 captions (an average of 3.6 moments per video), and is widely used for dense-event captioning and moment retrieval.

\subsubsection{Evaluation Metrics}
\label{sec: Evaluation Metrics.}

For fair comparison, we follow the evaluation protocol of prior work~\cite{Moment-GPT,vtimellm}, employing standard VMR metrics: R1@n, mAP@m, mAP@avg, and mIoU. Specifically, R1@n denotes the percentage of queries for which the top-1 prediction achieves an IoU greater than n with the ground truth. mAP@m refers to the mean average precision at IoU threshold m, while mAP@avg averages mAP across thresholds from 0.5 to 0.95 with a step size of 0.05. mIoU measures the average maximum IoU between predicted and ground-truth segments across all queries, indicating localization accuracy.
For Video Highlight Detection (VHD), we report HIT@1, reflecting whether the top-ranked segment matches the ground truth, and mAP, evaluating ranking quality.

\section{Instruction Pairs and Sensitivity Analysis}
\begin{figure*}[h!]
    \centerline{\includegraphics[width=1\textwidth]{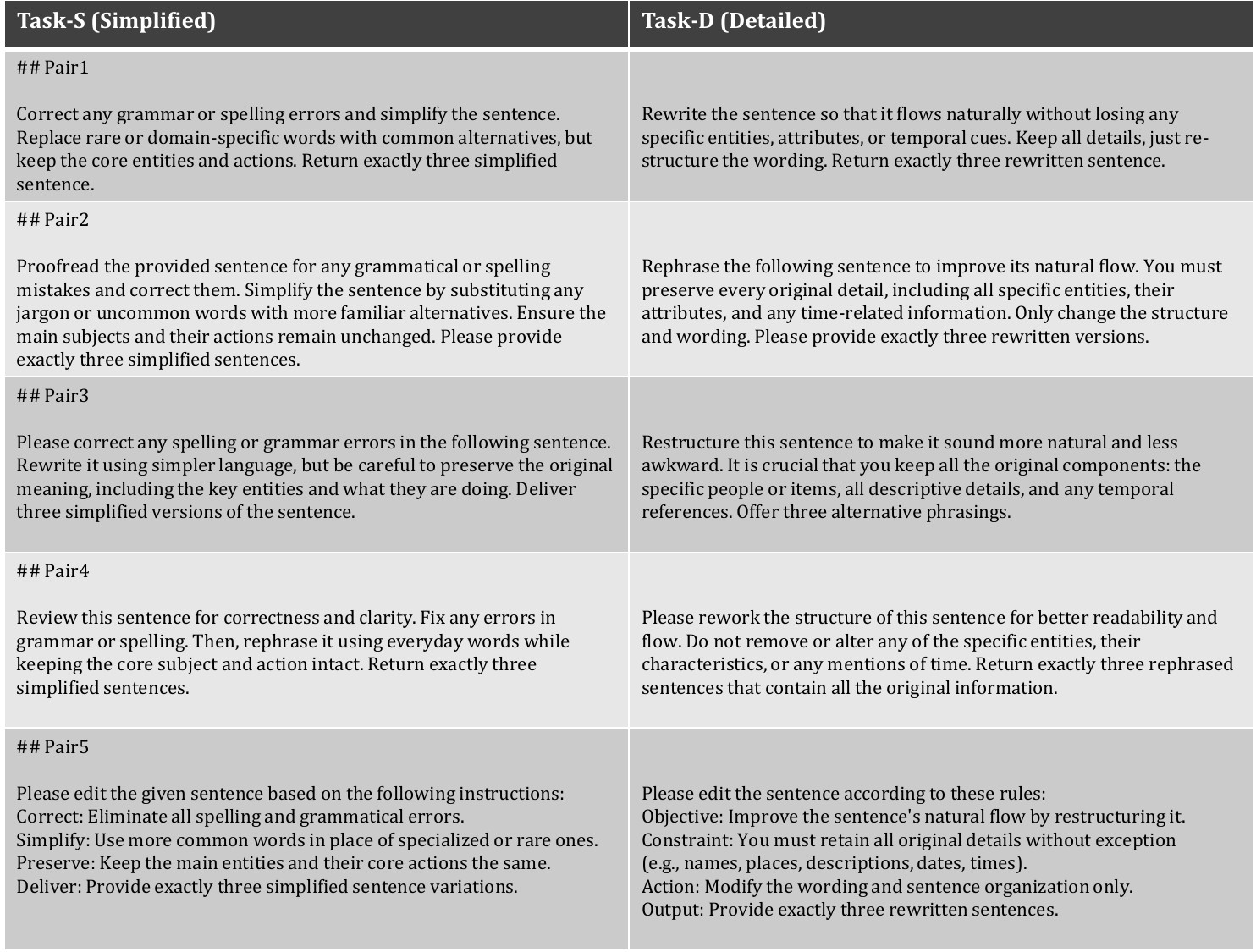}}
    \caption{Prompt for instruction.}
    \label{fig7}
\end{figure*}

This section provides the set of instruction pairs (shown in Figure~\ref{fig7}) and the performance for each pair for the granularity-based query rewriting described in Section 3.2. We present a sensitivity analysis to empirically validate our claim that the framework's performance is robust and not overly sensitive to the specific phrasing of the prompts.

To empirically verify the robustness of our approach, we evaluated the performance of each of the five instruction pairs on the QVHighlights validation set. 
The results in Table~\ref{tab:prompt_performance} demonstrate a high degree of performance consistency across the different phrasings. The R1@0.5 scores, for example, are all clustered within a tight 0.7\% range, from a minimum of 61.26\% to a peak of 61.94\%. This low variance provides strong empirical evidence that our framework has generalized the underlying concepts of 'simplification' and 'detail preservation' rather than overfitting to a single, fine-tuned prompt. This analysis validates the claim made in the main text that our granularity-based rewriting method is robust to variations in the instruction prompts. 
\begin{table}[h!]
\centering
\resizebox{0.9\columnwidth}{!}{%
\begin{tabular}{ccccc}
\toprule
\textbf{Pair} & \textbf{R1@0.5} & \textbf{R1@0.7} & \textbf{mAP@0.5} & \textbf{mAP@avg} \\
\midrule
\textbf{1} & \textbf{61.94} & \textbf{41.81} & 59.63 & \textbf{39.12} \\
2 & 61.81 & 41.60 & 59.27 & 39.01 \\
3 & 61.63 & 41.10 & 58.99 & 38.65 \\
4 & 61.52 & 41.35 &\textbf{ 59.78} & 38.93 \\
5 & 61.26 & 41.42 & 59.45 & 38.53 \\
\bottomrule
\end{tabular}}
\caption{Sensitivity analysis of the instruction pairs on the QVHighlights validation set. Consistent high performance is achieved across all five prompt pairs, demonstrating that our query rewriting methodology is robust and not overfit to a specific prompt phrasing.}
\label{tab:prompt_performance}
\end{table}
For our main experiments, we selected Pair 1 as the optimal combination as it achieved the best performance across most key metrics, including R1@0.5, R1@0.7, and mAP@avg.

\bibliography{aaai2026}

\end{document}